\documentclass[letterpaper]{article} 
\usepackage{template}  
\usepackage{times} 
\usepackage{helvet}  
\usepackage{courier}  
\usepackage[hyphens]{url} 
\usepackage{graphicx} 
\usepackage{natbib}  
\usepackage{caption} 
\DeclareCaptionStyle{ruled}{labelfont=normalfont,labelsep=colon,strut=off} 
\usepackage{algorithm}
\usepackage{algorithmic}
\usepackage{subcaption}
\usepackage{kotex}
\usepackage{multirow}
\usepackage{amsmath}
\usepackage{array}
\usepackage{amsfonts}
\usepackage[toc,page]{appendix}
\usepackage{siunitx}
\usepackage{newfloat}
\usepackage{listings}
\floatstyle{ruled}
\newfloat{listing}{tb}{lst}{}
\floatname{listing}{Listing}

\newcommand{\LSTM}{\ensuremath{\,\textrm{LSTM}}}

\title{Flexible Networks for Learning Physical Dynamics of Deformable Objects}
\author {
    Jinhyung Park, \textsuperscript{\rm 1}
    Dohae Lee, \textsuperscript{\rm 1}
    In-Kwon Lee \textsuperscript{\rm 1}
}
\affiliations {
    \textsuperscript{\rm 1} Yonsei University\\
    jh9604@yonsei.ac.kr, dlehgo1414@gmail.com, iklee@yonsei.ac.kr
}

\usepackage{bibentry}

\begin{document}

\maketitle

\begin{abstract}
Learning the physical dynamics of deformable objects with particle-based representation has been the objective of many computational models in machine learning. While several state-of-the-art models have achieved this objective in simulated environments, most existing models impose a precondition, such that the input is a sequence of ordered point sets. That is, the order of the points in each point set must be the same across the entire input sequence. This precondition restrains the model from generalizing to real-world data, which is considered to be a sequence of unordered point sets. In this paper, we propose a model named \emph{time-wise PointNet} (TP-Net) that solves this problem by directly consuming a sequence of unordered point sets to infer the future state of a deformable object with particle-based representation. Our model consists of a shared feature extractor that extracts global features from each input point set in parallel and a prediction network that aggregates and reasons on these features for future prediction. The key concept of our approach is that we use global features rather than local features to achieve invariance to input permutations and ensure the stability and scalability of our model. Experiments demonstrate that our model achieves state-of-the-art performance with real-time prediction speed in both synthetic dataset and real-world dataset. In addition, we provide quantitative and qualitative analysis on why our approach is more effective and efficient than existing approaches.
\end{abstract}

\section{Introduction}
Humans have the ability to reason on historical trajectories of various objects and easily make inferences on future states based on their intuitions. In order to mimic such ability, developing a computational model such as differentiable physics engine that can learn the physical dynamics of objects has been a core domain in computer vision, robotics, and in machine learning \cite{dfe2019robotics, psreview2021, endtoenddfe2018, brax2021paper}. To implement such computational models, physical reasoning on objects with particle-based representations has garnered attention in recent years \cite{macklin2014unified}.
As particle-based representations provide flexibility in representing objects with complex shapes and high degree of freedom, several studies have provided frameworks that can learn to simulate the physical dynamics of deformable objects with particle-based representation \cite{mrowca2018flexible, li2019learning, li2020visual}.

\begin{figure}[t]
    \centering
    \begin{subfigure}{0.45\textwidth}
        \centering
        \includegraphics[width=\textwidth]{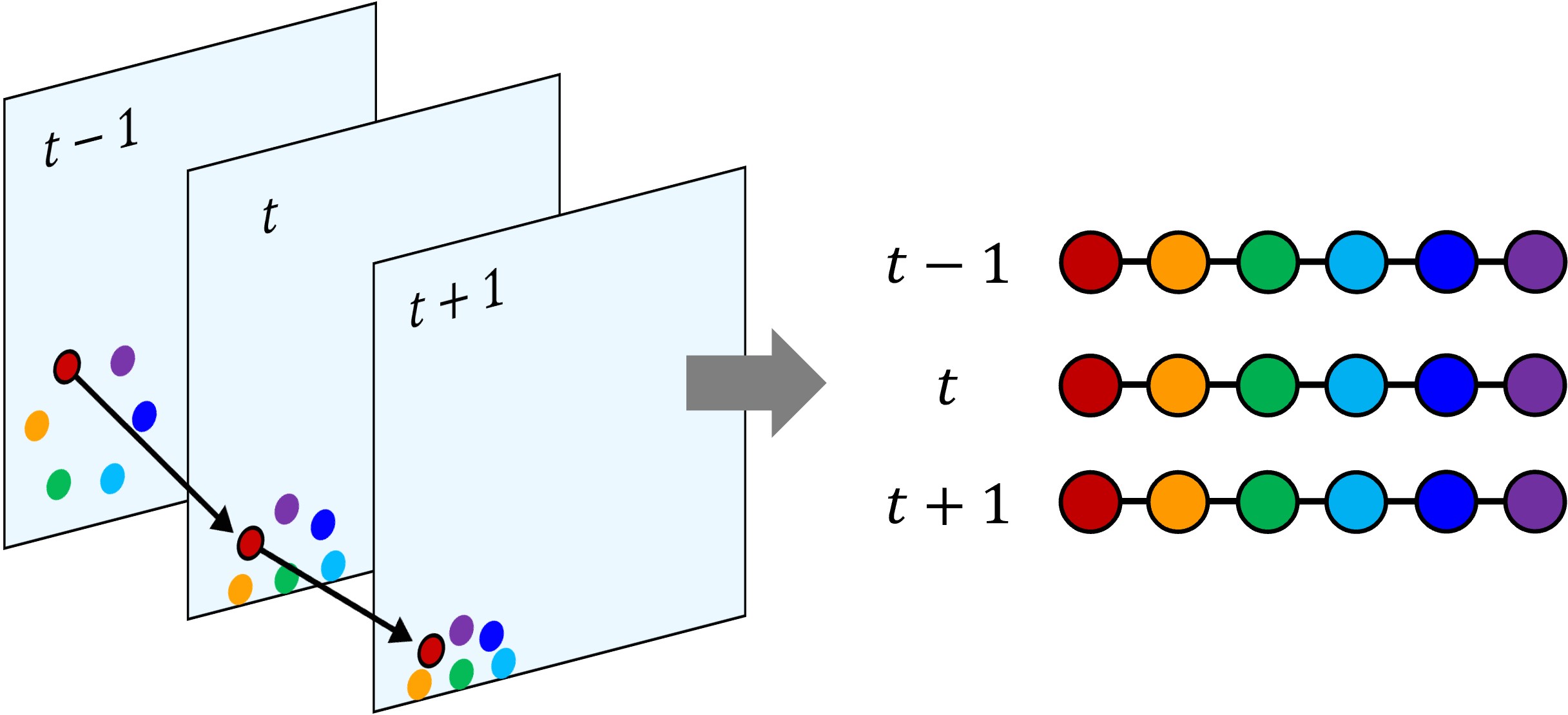}
        \caption{}
        \label{fig:ordered_pointset}
    \end{subfigure}
    \hfill
    \begin{subfigure}{0.45\textwidth}
        \centering
        \includegraphics[width=\textwidth]{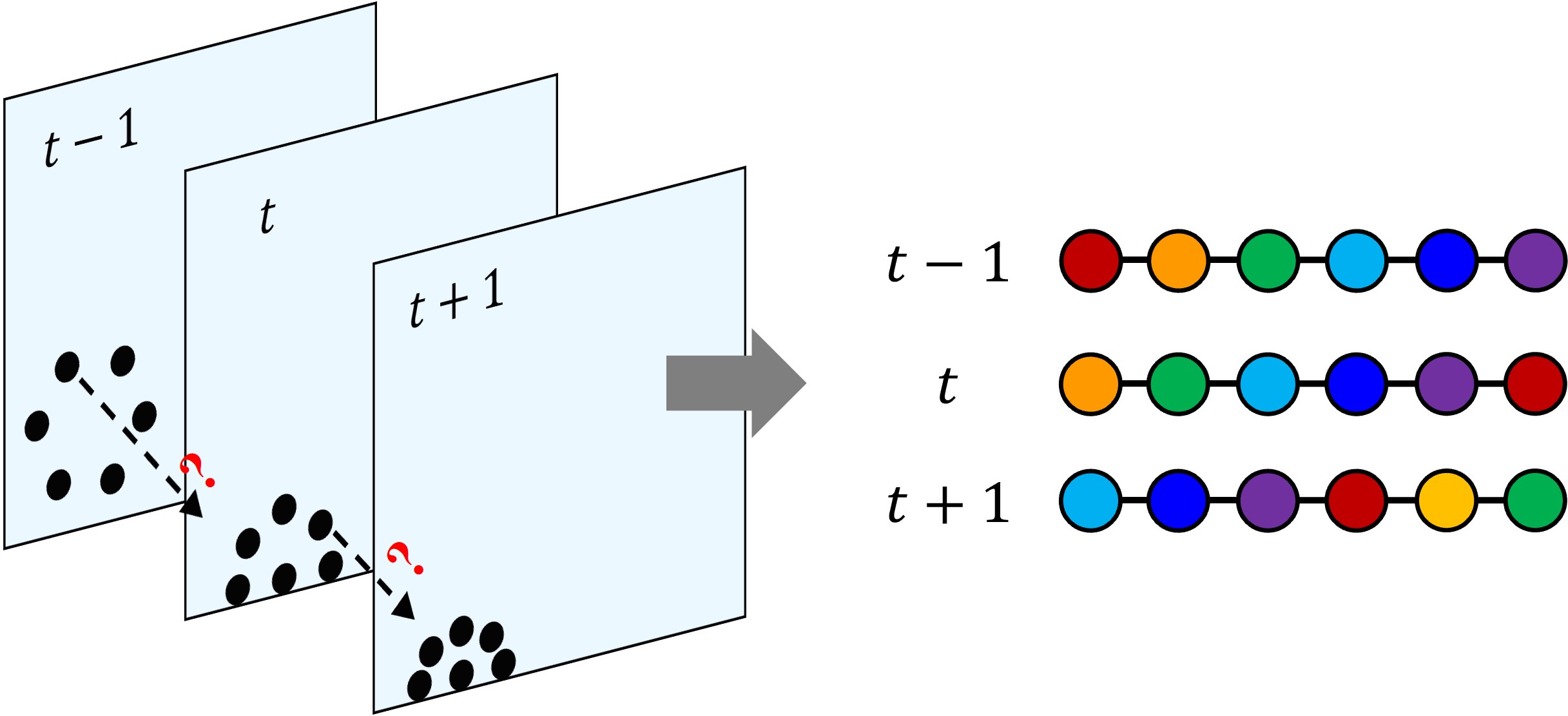}
        \caption{}
        \label{fig:orderless_pointset}
    \end{subfigure}
        \caption{(a) We can easily configure a sequence of ordered point sets in simulated environments. (b) The previous condition is seldom met in many real-world scenarios.}
        \label{fig:ordered_vs_orderless}
\end{figure}

\begin{figure*}[htbp]
\centering
\includegraphics[width=\textwidth]{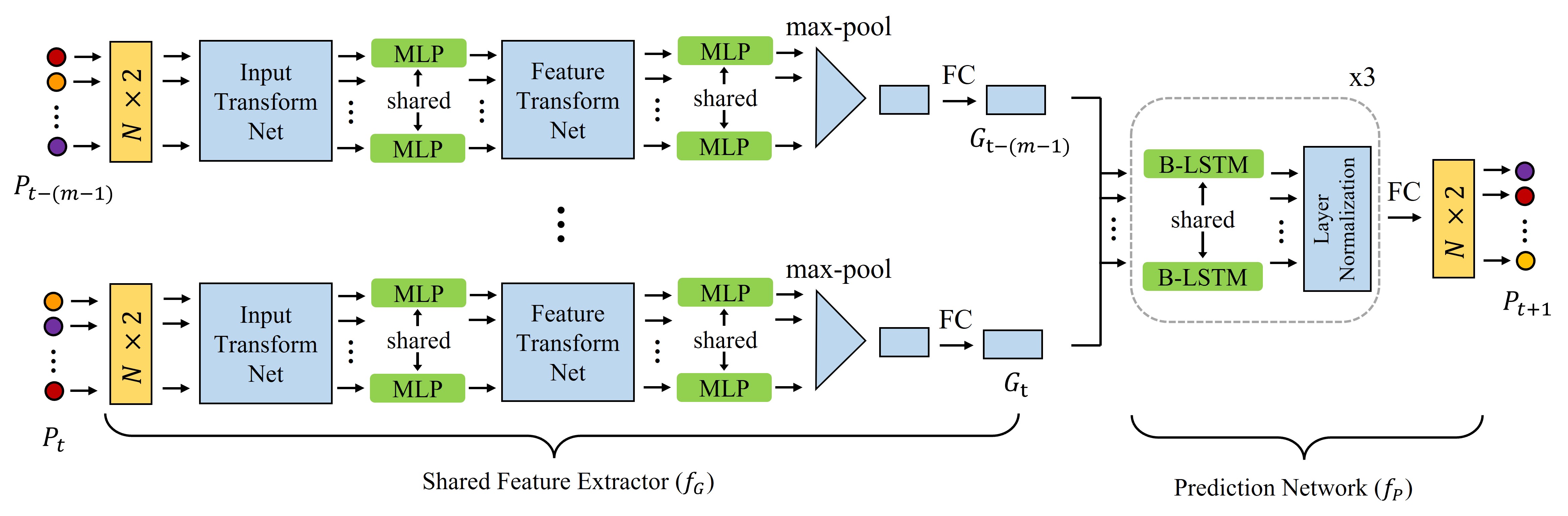}
\caption{TP-Net Architecture. The network takes a sequence of $m$ unordered point sets as input, and outputs the predicted unordered point set in the next time step. ``MLP", ``FC" and ``B-LSTM" stands for multi-layer perceptron, fully-connected layer and bidirectional-LSTM layer, respectively. The input transform net and feature transform net is directly adopted from the existing classification network -- PointNet.}
\label{fig:model_overview}
\end{figure*}

While these models have achieved state-of-the-art performances in predicting the future states of deformable objects, these models are optimized from learning \emph{a sequence of ordered point sets}. In other words, the order of the points in each point set must be the same across the entire input sequence when training or testing the model. One of the reasons for imposing this precondition in the data is to use a weight-shared LSTM \cite{hochreiter1997long}  layer in the model architecture or to easily reason on the historical trajectories of individual points. However, this approach is only feasible in simulated environments where we can map each point in the current frame to its corresponding point in the previous frame (Fig. \ref{fig:ordered_vs_orderless} (a)). In many real-world scenarios, individual point correspondences between different frames are not observable, thus making it challenging to configure a sequence of ordered point sets from the given input (Fig. \ref{fig:ordered_vs_orderless} (b)).
Although sorting the points in each point set into a canonical order might seem to resolve this problem, PointNet \cite{pointnet} demonstrates that in high-dimensional space, there exists no ordering that is stable with respect to point perturbations.
Therefore, sorting-based approaches also result in lower performance which we will show this empirically in Section IV.

In order to build a flexible neural network for constructing a differentiable physics engine that is independent of this concern, we propose our model named \emph{time-wise PointNet} (TP-Net) that can directly consume a sequence of unordered point sets to infer the future dynamics of a deformable object. TP-Net extends a key concept of an existing classification network called PointNet \cite{pointnet}, and achieves invariance to $(N!)^m$ input permutations, where $N$ denotes the number of points in each input point set and $m$ denotes the length of the input sequence, i.e., the number of point sets in the sequence that the model takes as input. In Section \ref{sbsec:method_section}, we demonstrate how our model achieves this property and explain why extending a key idea from a classification network to design a time-series prediction model for physical reasoning is an effective approach. The results from our experiments indicate that TP-Net outperforms existing models, including state-of-the-art, in predicting the future states of both simulated and real-world deformable objects, with $O(N)$ time and space complexity.

The key contributions of our work are as follows:
\begin{itemize}
    \item We design a deep neural network that can directly consume a sequence of unordered point sets to predict the future particle positions of simulated and real-world deformable objects;
    \item Based on qualitative and quantitative analysis, we show how the key idea from an existing classification network can be extended to develop an effective time-series prediction model for physical reasoning;
    \item We demonstrate the efficiency of our approach compared to existing approaches for consuming a sequence of unordered point sets, based on theoretical and empirical analysis.
\end{itemize}

\section{Related Work}
\subsubsection{Physical Reasoning on N-Body System}
An early attempt to develop a learnable and differentiable physics engine that can reason on interacting objects was achieved by the interaction network \cite{battaglia2016interaction}. The interaction network showed the effectiveness of utilizing graph neural networks (GNN) \cite{scarselli2008graph} for physical reasoning on systems with $N$ bodies, which inspired several other studies to propose learnable simulators based on GNN \cite{cranmer2020discovering, kipf2018neural, dNRI, fNRI, SUGAR, Martinkus_Lucchi_Perraudin_2021, DSLR2021, gnnreview2021}. These approaches have in common that for each time step, the system is represented as a graph where each node and edge represents an object and the relationship between objects, respectively. By updating the state of each node over time with message-passing \cite{duvenaud2015convolutional}, the network can learn to infer the future states of each node, based on relational reasoning \cite{relationalReasoning} and relational inductive bias \cite{relationalInductiveBias}. 

Recently, dNRI \cite{dNRI} and NRI \cite{kipf2018neural} have demonstrated its effectiveness in using GNN to predict the interactions between particles purely from observational data. These approaches are also applicable in predicting the future state of a deformable object that is modeled with mass-spring system. However, these models can only learn to predict the future from a sequence of ordered point sets. Although a graph for each time step can be constructed from a \emph{single} unordered point set since a graph is invariant to input permutations of the nodes, the order of the points matters when reasoning on two or more different point sets. For example, if the input point set at time step $t$ and $t+1$ are given as $\{(x_1^t, y_1^t), (x_2^t, y_2^t), (x_3^t, y_3^t)\}$ and $\{(x_2^{t+1}, y_2^{t+1}), (x_1^{t+1}, y_1^{t+1}), (x_3^{t+1}, y_3^{t+1})\}$, respectively, graph based networks learn a message-passing function based on the assumption that $(x_1, y_1)$, $(x_2, y_2)$, and $(x_3, y_3)$ changes to $(x_2^{t+1}, y_2^{t+1})$, $(x_1^{t+1}, y_1^{t+1})$, and $(x_3^{t+1}, y_3^{t+1})$, respectively. Therefore, training or testing the network with misaligned point set sequences results in poor performances, thus restraining the model to generalize to real-world data where a sequence of ordered point sets is not configurable. We empirically prove such behavior in Section \ref{sec:experiments_real_world_dataset}.

\subsubsection{Physical Reasoning on Deformable Objects}
Similar approaches of utilizing GNN were adopted to develop a learnable simulator for deformable objects \cite{mrowca2018flexible, li2019learning, li2020visual, sanchez2020learning}. In these studies, deformable objects are modeled with particle-based representation to make future predictions on individual particles, based on the learned dynamics of the system. These approaches showed excellent performances in inferring the future dynamics of simulated deformable objects, yet they assume a few conditions that are seldom met in practice. In addition to the assumption that these approaches assume a sequence of \emph{ordered} point sets as their input, the hierarchical relation network \cite{mrowca2018flexible} and dynamics prior \cite{li2019learning, li2020visual} also require additional attributes as input, such as relational attributes that describes the relationship between individual particles in the system. Since relational attributes are not observable in many real-world scenarios, the model lacks generalization ability and shows poor performance without those attributes, which we show this behavior in Section \ref{sec:experiment}.

\subsubsection{Consuming Sequence of Unordered Point Sets}
\label{sec:consuming_unordered}
While our method of consuming a sequence of unordered point sets is fundamentally different from existing approaches, a few studies have also proposed networks that can process a sequence of unordered point sets for various purposes \cite{min_CVPR2020_PointLSTM, fan19pointrnn, gomes2021spatiotemporal, liu2019meteornet}. The key idea common in these approaches is that for each point in the current frame, the network searches for k-nearest-neighbors or relevant points in the previous frame. By combining state information of neighboring points in the past, the model achieves permutation invariance while preserving the spatial structure of the input. While this approach showed state-of-the-art performance in gesture recognition, such method shows poor performance when applied in future prediction (see results of GraphRNN in Section \ref{sec:experiments_sim_dataset}). As the network significantly depends on local features and neighbor points of each point, it suffers from serious performance degradation when outliers or unseen points exist in the input, which is a common scenario in long-term prediction. In addition, finding spatio-temporal neighbor points for every point in the point set is a $O(N^2)$ operation -- quadratic to the number of input points -- that is computationally expensive and lacks scalability to systems with large number of particles.

\section{Method}
\label{sbsec:method_section}
We define an unordered point set with $N$ points at time $t$ as $P_t = \{ {p_{1, t}}, {p_{2, t}}, ... {p_{N, t}}\}$ with $p_{i, t} \in \mathbb{R}^2$ being the Euclidean coordinates of point $i$ in $P_t$. Then, we formulate our problem by learning a continuous function $f$ that infers the future point set $P_{t+1}$ of a deformable object modeled with mass-spring system, when a sequence of $m$ unordered point sets is given as input:
    \begin{equation}
        P_{t+1} = f(P_{t-(m-1)}, \ P_{t-(m-2)}, \ ... \ , P_{t-1}, \ P_t).
    \end{equation}

\subsection{Model Overview}
\label{sbsec:model_architecture}
To develop a framework that can process a sequence of unordered point sets for physical reasoning, we propose our model called TP-Net, which has two key modules: shared feature extractor and prediction network (Fig. \ref{fig:model_overview}). The key idea of our model is that we aggregate global features of each point set to predict individual particle positions in the next time step. In addition to making TP-Net invariant to $(N!)^m$ input permutations, this approach is also robust to input perturbations and outliers, thereby facilitating stable long-term predictions, especially in predicting the shape deformation of the object. We explain how TP-Net achieves these objectives in the following section and demonstrate its effectiveness in Section \ref{sec:experiment}.

\subsection{Shared Feature Extractor}
\label{sbsec:shared_feature_extractor}
We first design our shared feature extractor that learns a continuous set function $f_G$, which takes a single unordered point set $P_t$ as input and outputs global features $G_t$ as described in \eqref{shared_feature_extractor_equation}. Then, we apply this to each one of $m$ input point sets in parallel. This process can be represented by the following equations:
    \begin{equation}
        G_{t-(m-1)} = f_G(P_{t-(m-1)}),
    \end{equation}
    \begin{equation*}
        ... \\
    \end{equation*}
    \begin{equation}
        G_{t-1} = f_G(P_{t-1}), \\ 
    \end{equation}
    \begin{equation}
    \label{shared_feature_extractor_equation}
        G_{t} = f_G(P_{t}).
    \end{equation}

Since the global features extracted from each input point set are invariant to $N!$ input permutations (i.e. invariant to the order of the points in the input point set), we achieve invariance to $(N!)^m$ input permutations in total since we extract global features from $m$ input point sets independently. While global features $G_t$ are invariant to the order of the points in the input point \emph{set}, we define global features $G_t$ as a \emph{sequence} of features:
\begin{equation}
\label{global_f_eq}
    G_t = (g_{1}^t, \ g_2^t, ... \ , g^t_k),
\end{equation}
where $g_i^t$ is the $i$-th feature of the point set $P_t$, and $k$ is the number of features we encode from a point set. The purpose of defining global features as a \emph{sequence} of features rather than a \emph{set} of features, is to align the same type of features of each point set when we aggregate multiple global features $G_{t-(m-1)}, \ ... \ G_{t-1}, \ G_t$ in the prediction network. This allows the prediction network to apply a weight-shared LSTM layer on these aligned global features.

\subsubsection{Implementation Details}
Our shared feature extractor adopts a submodule of the classification network from PointNet \cite{pointnet}. The classification network from PointNet takes a single unordered point set as input and outputs the predicted score of each class for object classification. While we use the same structure of this network to design our shared feature extractor, we modify the last part of the network to output global features. We also adjust the input vector shape and layer sizes to process 2D data, instead of 3D data, since we use 2D datasets (Fig. \ref{fig:model_overview}).

The network first applies a series of shared MLP layers on each point independently for input transformations, feature transformations, and local features extraction. Then, the network aggregates per point features by max pooling to extract global features of the point set. Although the original PointNet applies another series of MLP layers and a softmax function on these global features to compute the scores for object classification, our network applies a single MLP layer and a ReLU function on these global features instead, to refine and extract the final global features as output.

\subsubsection{Extending PointNet for Future Prediction}
\label{sec:extension_to_future_prediction}
Although PointNet is originally designed to capture relevant features for classification or semantic segmentation tasks, PointNet is also an excellent source for future frame prediction of deformable objects with particle-based representation. PointNet learns a continuous set function that summarizes the shape of an object into a sparse set of key points. Since the global features extracted from PointNet are bottleneck features of these key points, these features are robust to small perturbation or outliers in the input. Therefore, we extract global features from each point set in parallel and predict the future particle positions by reasoning on these global features sequentially, with the expectation that our model will be robust to outliers during prediction rollout. In Section \ref{sec:experiment}, we show this behavior of TP-Net and prove its effectiveness.

\subsection{Prediction Network}
The prediction network learns a continuous function $f_P$, that takes global features of each input point set as input and predicts the future particle positions $P_{t+1}$:

    \begin{equation}
        P_{t+1} = f_P(G_{t-(m-1)}, \ ... \ , \ G_{t-1}, G_t). \\ \\
    \end{equation}

\subsubsection{Implementation Details}
We implement our prediction network by first aggregating the global features $G_{t-(m-1)}$, $...$, $G_{t-1}$, $G_t$ into a sequence of global features $\mathcal{G}$:
    \begin{equation}
     \mathcal{G} = (G_{t-(m-1)}, \ ... \ , \ G_{t-1}, \ G_t), \\
    \end{equation}
to preserve the temporal structure of the original input. Note that global features per every time step are also a \emph{sequence} of features as illustrated in Equation (\ref{global_f_eq}). Since we apply the same shared feature extractor to all input point set in parallel, the order of the features in every global features is the same -- i.e., $G_i = (g_{1}^i, \ g_2^i, ... \ , g_k^i)$ for all $i = t-(m-1), \ ... \ , t-1, \ t$. Since $\mathcal{G}$ maintains the same order of features for every $G_i$ across its entire sequence, we then apply a bidirectional LSTM layer to $\mathcal{G}$ with the assumption that each feature $g_j^t$ has an independent hidden state $h_j^t$ and a cell state $c_j^t$:
    \begin{equation}
        h_j^t, \ c_j^t = \LSTM \ (g_j^t, \ h_j^t, \ c_j^t). \\
    \end{equation}
After we apply three bidirectional LSTM layers (with layer normalization \cite{ba2016layer} in between bidirectional LSTM layers), we apply a fully-connected layer and reshape its output to finally predict the 2D point set $P_{t+1}$ (Fig. \ref{fig:model_overview}).

\subsection{Model Training}
\subsubsection{Dataset}
We first train and test our model on a synthetic dataset of moving deformable object with particle-based representation. Then, we directly test our model on real-world dataset that is similar to the synthetic dataset in terms of object radius, velocity and rigidness, but essentially different in that the order of the particles in each point set is unordered.

For synthetic dataset, we use the Box2D physics engine \cite{catto2020box} to generate 7125 simulated trajectories of a deformable object. We model our deformable object with a mass-spring system where 30 particles comprises the boundary of the object and each particle is jointly connected to a number of other particles by a restorable spring (Fig. \ref{fig:softbody}). When generating the simulated trajectories, we randomize the initial position, initial force and initial direction of the object.

For real-world dataset, we capture 40 trajectories of a real-world deformable object by using a high speed camera. Then, we preprocess the video data to construct a sequence of unordered point sets that represents the trajectory of the object. For details on how we prepare and preprocess our synthetic and real-world dataset, please refer to Section \ref{sec:data_generation} in the appendix.

\subsubsection{Loss Function and Hyperparameters}
We define our loss function $L_{CD}$ by using the Chamfer distance \cite{barrow1977parametric} between the ground truth point set $P$ and the predicted point set $\hat{P}$:
    \begin{equation}
        L_{CD}(P, \hat{P}) = \sum^{}_{x\in P}{\min^{}_{y \in \hat{P}} \ \lVert x-y \rVert^{2}_{2}} + \sum^{}_{y\in \hat{P}} {\min^{}_{x \in P} \ \lVert x-y \rVert^{2}_{2}}.
    \end{equation}
In order to empower our model to achieve long-term prediction ability with robustness to input perturbations, we make our model to recursively predict eight future point sets during training. Thus, we define our final loss $L$ as the sum of the losses of each predicted point set:
    \begin{equation}
        L = \sum^{8}_{i=1}{w_{i}L_{CD}(P_{t+i}, \hat{P}_{t+i})},
    \end{equation}
where $w_1=9$ and $w_i=1$ for $i=2, \ 3, \ ... \ , 8$ to balance the weight of predicting future point sets between when a clean input is given and when input with noise is given.

When training our model, we use a learning rate of $10^{-3}$ with the Adam optimizer \cite{adam} and train it for 100 epochs with a batch size of 128 on Intel(R) Core(TM) i5-8400 CPU @ 2.80Hz and GeForce GTX1080 Ti for 18 to 24 hours. \\

\section{Experiment}
\label{sec:experiment}

\begin{figure*}[t]
\centering
\includegraphics[width=\textwidth]{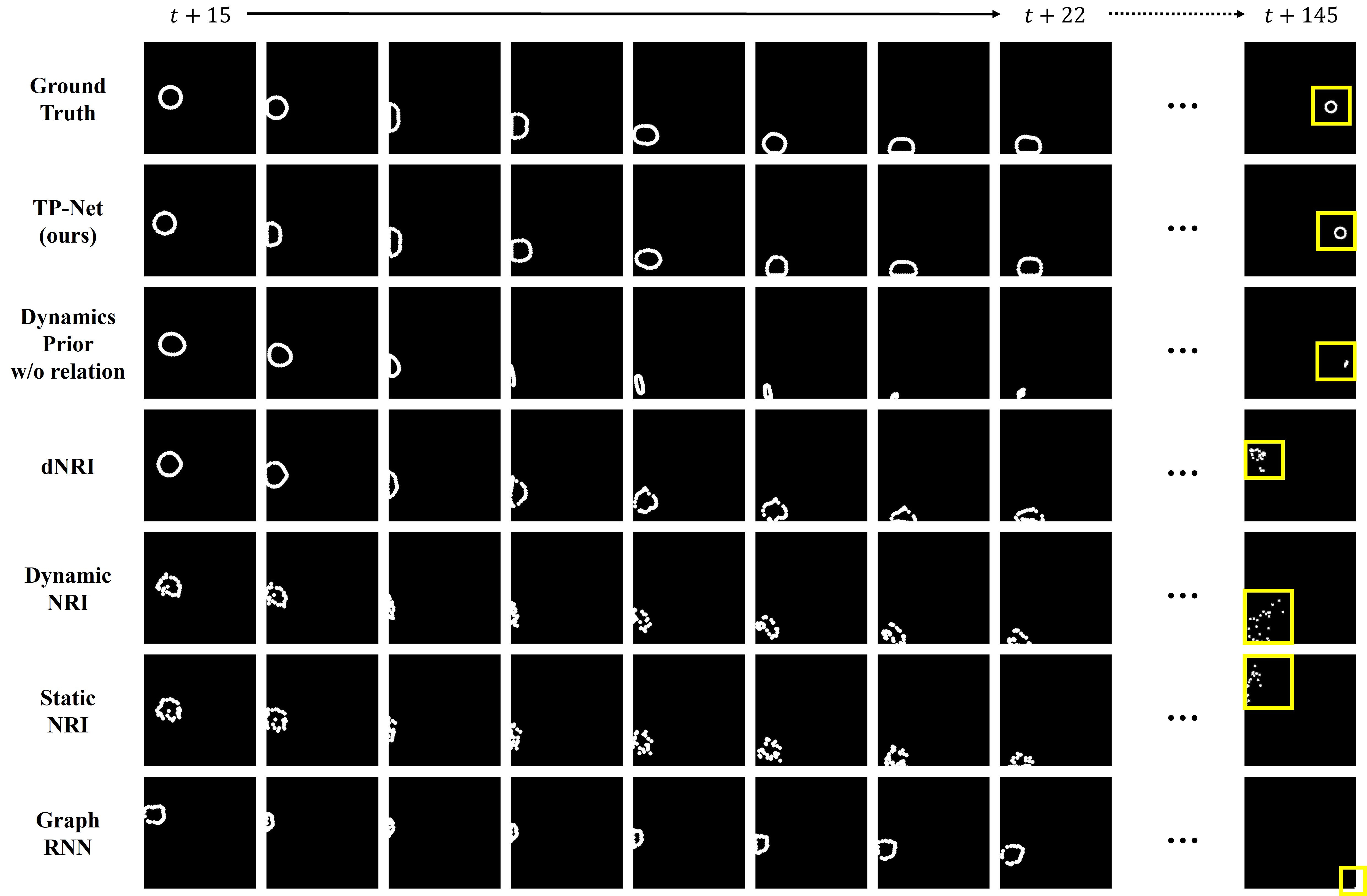}
\caption{Future prediction on moving deformable object (synthetic dataset). We visualize the prediction results after 15 to 22 time steps and after 145 time steps when five input frames are given initially. For the eight output frames from $t+15$ to $t+22$, we zoom-in the left-bottom quadrant of the frame. For the output frame at $t+145$, we draw a yellow bounding box to locate the deformable object in the frame.}
\label{fig:synthetic_result}
\end{figure*}

\begin{table*}[t]
\renewcommand{\arraystretch}{1.2}
\centering
\caption{Average rollout error of future prediction on moving deformable object (synthetic dataset). CD indicates the position error (Chamfer distance) and CD* indicates the shape error (variation of the vanilla Chamfer distance) of the prediction made from each model. The asteroid(*) next to DynamicsPrior denotes that the model is trained and tested without relation attributes.}
\label{synthetic_quantitative}
\begin{small}
\begin{tabular}{l|c|c|c|c|c|c|c|c|c|c|c|c}
\hline\hline
& \multicolumn{4}{c|}{\# input frames = 3} & \multicolumn{4}{c|}{\# input frames = 4} & \multicolumn{4}{c}{\# input frames = 5} \\
\hline
Prediction Steps & \multicolumn{2}{c|}{40} & \multicolumn{2}{c|}{80} & \multicolumn{2}{c|}{40} & \multicolumn{2}{c|}{80} & \multicolumn{2}{c|}{40} & \multicolumn{2}{c}{80} \\

\hline
    Rollout Error & CD & CD* & CD & CD* & CD & CD* & CD & CD* & CD & CD* & CD & CD* \\
\hline
 TP-Net (ours) & \textbf{0.094} & \textbf{0.007} & \textbf{0.246} & \textbf{0.007} & \textbf{0.071} & \textbf{0.007} & \textbf{0.164} & \textbf{0.007} & \textbf{0.101} & \textbf{0.008} & \textbf{0.225} & \textbf{0.007} \\
 DynamicsPrior* & 0.110 & 0.044 & 0.252 & 0.050 & 0.074 & 0.040 & 0.186 & 0.046 & 0.116 & 0.041 & 0.231 & 0.064 \\
 dNRI & 0.291 & 0.031 & 0.867 & 0.055 & 0.092 & 0.024 & 0.882 & 0.046 & 0.106 & 0.022 & 0.613 & 0.035 \\
 DynamicNRI & 0.788 & 0.113 & 0.760 & 0.292 & 0.313 & 0.036 & 0.946 & 0.047 & 0.254 & 0.031 & 0.882 & 0.044 \\
 StaticNRI & 0.668 & 0.143 & 0.714 & 0.286 & 0.312 & 0.036 & 0.946 & 0.047 & 0.254 & 0.031 & 0.882 & 0.044 \\
 GraphRNN & 1.235 & 0.055 & 2.398 & 0.069 & 0.370 & 0.030 & 0.699 & 0.054 & 0.342 & 0.024 & 1.154 & 0.092 \\
 
\hline
\hline
\end{tabular}
\end{small}
\end{table*}

We study our framework using two datasets: synthetic dataset and real-world dataset. For each dataset, we quantiatively and qualitatively compare the performance of TP-Net with other benchmarks -- dNRI \cite{dNRI}, StaticNRI \cite{kipf2018neural}, DynamicNRI \cite{kipf2018neural}, DynamicsPrior \cite{li2019learning, li2020visual}, and GraphRNN \cite{gomes2021spatiotemporal} - that are capable of predicting the future states of a deformable object. Then, we analyze and compare the time and space complexity of our model with GraphRNN.

\subsubsection{Evaluation Metrics}

For all experiments, we quantitatively measure the rollout error in two aspects: position error and shape error. We first define the position error $E_{p}$ as the average Chamfer distance between the ground truth point set $P$ and the predicted point set $\hat{P}$:
\begin{small}
    \begin{equation}
    \label{position_error_equation}
        E_{p}(P, \hat{P}) = \frac{1}{|P|} \left( \sum^{}_{x\in P}{\min^{}_{y \in \hat{P}} \ \lVert x-y \rVert^{2}_{2}} + \sum^{}_{y\in \hat{P}}{\min^{}_{x \in P} \ \lVert x-y \rVert^{2}_{2}} \right ).
    \end{equation}
\end{small}
However, the average Chamfer distance cannot fully reflect the shape difference between two point sets, as it simply measures the positional difference between individual particles (refer to Section \ref{sec:adoption_of_shape_error} in the appendix for details). Therefore, we define the shape error $E_s$ by using a variation of the vanilla Chamfer distance. We first apply a different translation matrix to each point set $P$ and $\hat{P}$, such that the center of each point set can be translated to the origin. Then, we compute the average Chamfer distance on these translated point sets. The equation for computing the shape error is as follows:
    \begin{equation}
    \label{shape_error_equation}
        E_{s}(P, \hat{P}) = E_{p}(M_{c}P, \ M_{\hat{c}}\hat{P}), \\
    \end{equation}
where $c$ is the center of mass of $P$, $\hat{c}$ is the center of mass of $\hat{P}$, $M_c$ is a translation matrix that translates $x$ to $x - c_x$ and $y$ to $y - c_y$, and $M_{\hat{c}}$ is a translation matrix that translates $x$ to $x - \hat{c}_x$ and $y$ to $y - \hat{c}$. In this way, we directly compare the geometric shape of two point sets regardless of their relative positions in the Euclidean space.

\subsubsection{Environments}
We train each model on synthetic dataset three times by giving different input conditions: giving three, four, and five input frames. Although our model TP-Net and GraphRNN can process both sequences of unordered and ordered point sets, we train and evaluate each model by solely using sequences of ordered point sets from the synthetic dataset to respect the precondition of the input for dNRI, StaticNRI, DynamicNRI, and DynamicsPrior. Moreover, we only use the position data for the input when training and evaluating each model, since other attributes are difficult to obtain from real-world data. Therefore, we zero-pad the relation attribute in the input when training and testing DynamicsPrior. Also, we use the positional difference along the time steps as the velocity attribute when training dNRI, StaticNRI, and DynamicNRI.

\subsection{Synthetic Dataset}
\label{sec:experiments_sim_dataset}
\subsubsection{Qualitative Comparison}
We test each model and visualize the predicted points along with the ground truth for qualitative comparison. As illustrated in Fig. \ref{fig:synthetic_result}, TP-Net is the only model that well predicts both the future trajectory and shape deformation of the object. Generally, all other models except TP-Net show significant performance degradation after the first or second collision of the object. While DynamicsPrior can predict the future trajectory of the object with high accuracy even in long term prediction, TP-Net significantly outperforms DynamicsPrior in predicting the shape deformation. This proves the effectiveness of our approach -- using max-pooled global features to predict the future state of each particles. Although TP-Net intermittently generates an outlier during the rollout, the model selects informative features from the point set and corrects those outliers in the next time step, thus maintaining reasonable shape of the object in long term. Conversely, model that uses neighboring points or local structures to predict the future state of each points, e.g., GraphRNN, shows significantly poor performance in long-term prediction.

\subsubsection{Quantitative Comparison}
We also evaluate and compare the performance by using quantitative metrics. We compute the average rollout error on 60 simulated trajectories of a deformable object, and compare its value after 40 and 80 time steps, respectively. As presented in Table \ref{synthetic_quantitative}, our model TP-Net achieves state-of-the-art performance in terms of both metrics -- position error and shape error. Although DynamicsPrior shows as high performance as TP-Net in terms of position error, the metric needs additional interpretation. As illustrated in Fig. \ref{fig:synthetic_result}, DynamicsPrior is simply good at predicting the general trajectory of individual particles without reasoning on its geometric shape, which could still result in low position error. In fact, TP-Net not only outperforms DynamicsPrior by a large margin (6-9 times) in terms of shape error, but also maintains a relatively constant shape error even in long term prediction.

Moreover, note that the analysis above is based on the results when each model is trained and tested solely on sequences of ordered point sets. It is clear that training and testing each model on sequences of unordered point sets will result in poor performances for dNRI, StaticNRI, DynamicNRI, and DynamicsPrior, since they highly depend on the order of the points in each input point set. We will also show this behavior in the next section.

\begin{figure*}[ht]
\centering
\includegraphics[width=\textwidth]{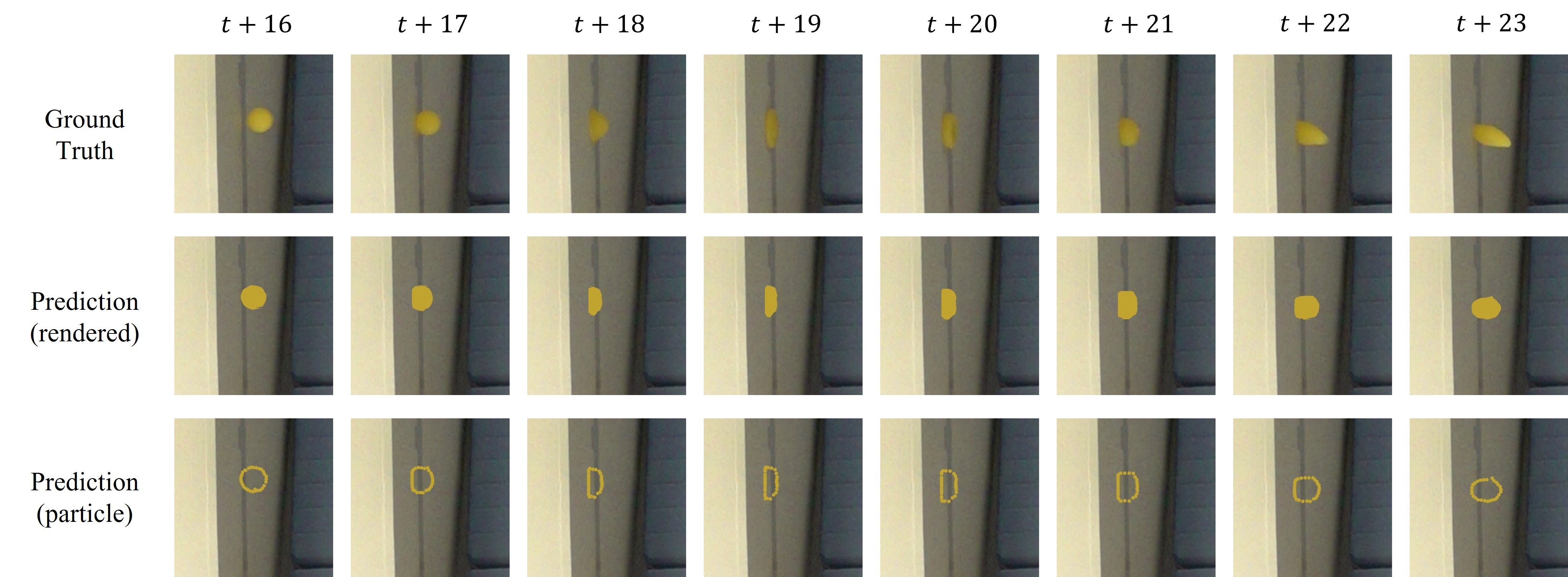}
\caption{\label{fig:real_world_result}Future predictions on real deformable object. The images in the first row are the original frames from the camera. The images in the bottom row are generated by plotting the predicted positions of each particle in the background image. The images in the middle row are generated by painting the interior of the contour drawn by connecting the predicted particles.}
\end{figure*}

\begin{figure*}[ht]
    \centering
    \begin{subfigure}{0.49\textwidth}
        \centering
        \includegraphics[width=1.0\textwidth]{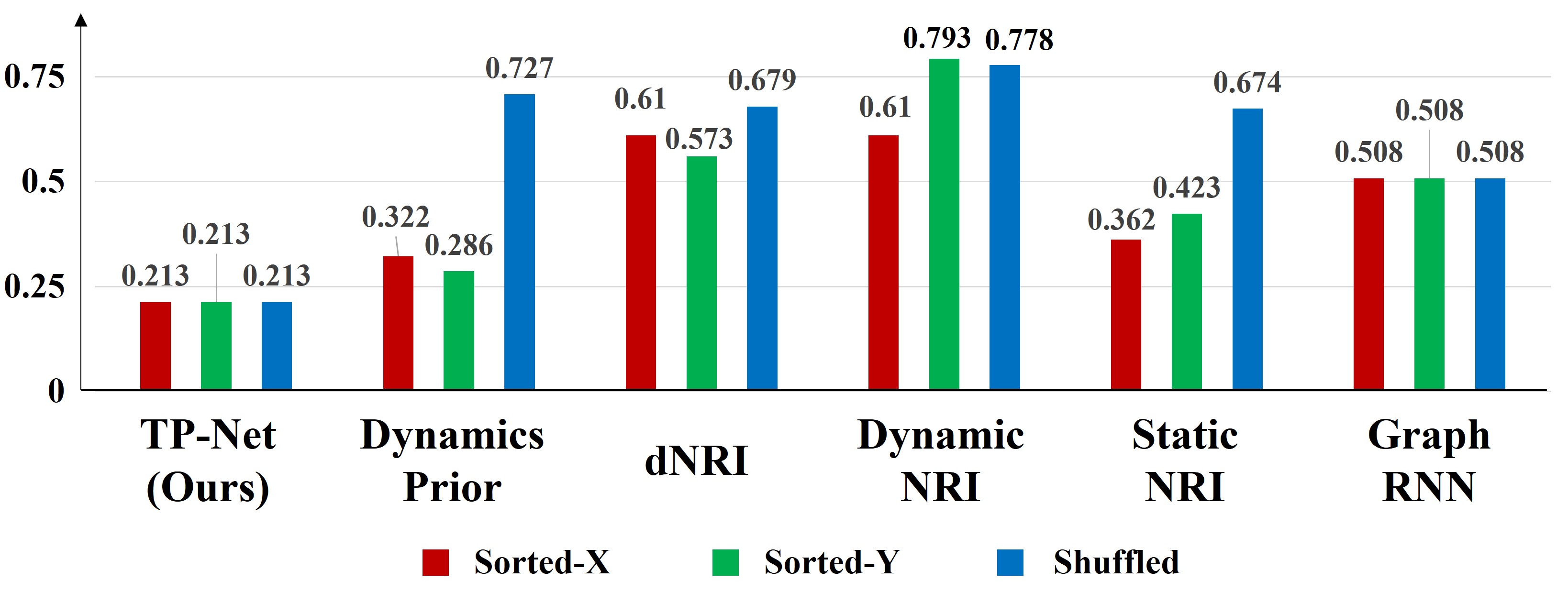}
        \caption{Average position error of future prediction on real-world data}
        \label{fig:real_world_quantitative_1}
    \end{subfigure}
    \hfill
    \begin{subfigure}{0.49\textwidth}
        \centering
        \includegraphics[width=1.0\textwidth]{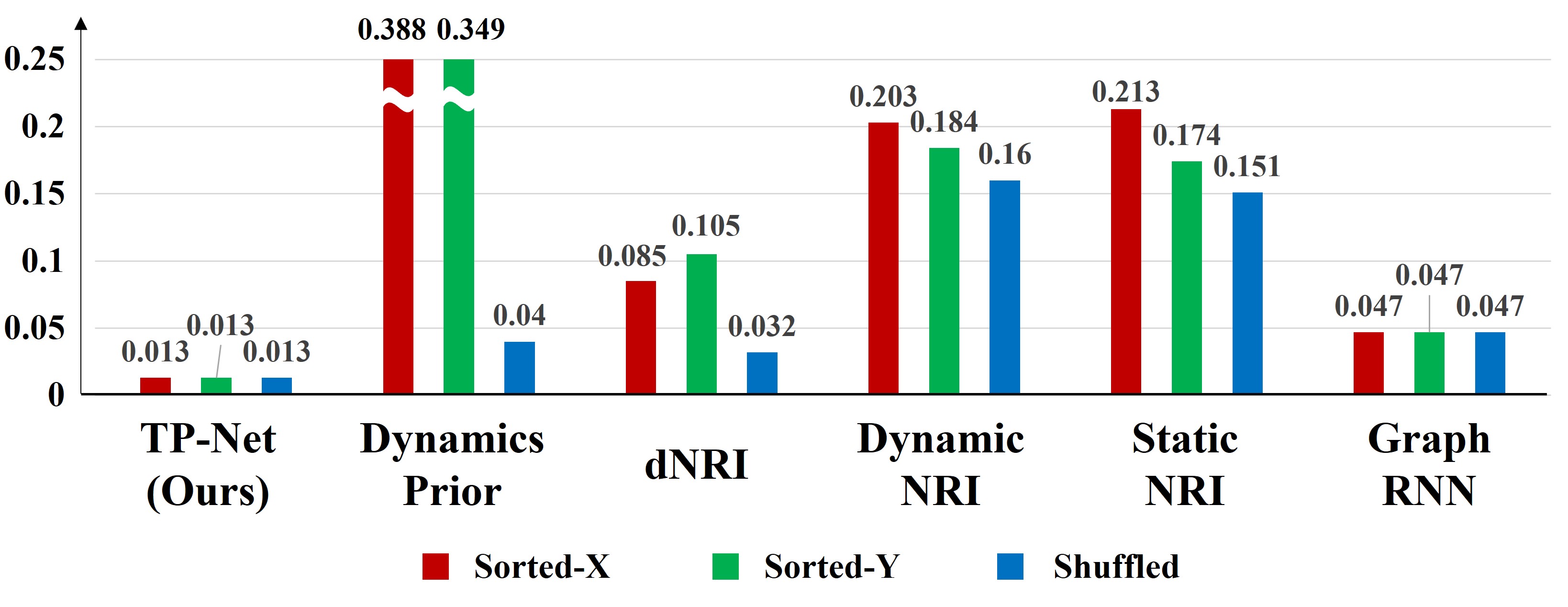}
        \caption{Average shape error of future prediction on real-world data}
        \label{fig:real_world_quantitative_2}
    \end{subfigure}
        \caption{For each model, we measure the rollout error after 40 time steps of each scenario with different input ordering method.}
        \label{fig:real_world_quantitative}
\end{figure*}

\subsection{Real world Dataset}
\label{sec:experiments_real_world_dataset}
\subsubsection{Generalization to Real World Data}
To demonstrate the generalization ability of TP-Net to real-world scenarios, we test our model that was only trained on synthetic dataset, directly on real-world dataset. Since we cannot configure a sequence of ordered point sets for input from real-world data, we arbitrarily configure the order of the points in each input frame and validate our model's ability in processing a sequence of unordered point sets. We give five steps of particle positions as input and make our model predict the future particle positions until the object disappears from the scene. 

The results illustrated in Fig. \ref{fig:real_world_result} shows that the learned policy of our model shows great potential to generalize to real world scenarios. Without knowing any information of the point correspondences between different input frames, TP-Net can reason on the underlying physical dynamics and infer the future states of a real world deformable object with reasonable accuracy.

\subsubsection{Comparison to Other Works}
We also compare the generalization ability of TP-Net to other benchmarks. To test the robustness of each model to different input ordering methods, we test each model on each test case with three different kinds of input. The three different input ordering method we use to preprocess the input are as follows: (1) Sorting the points in each input frame in ascending $x$ positions, (2) sorting the points in each input frame in descending $y$ positions, and (3) randomly shuffling the order of the points in each input frame.

We evaluate the rollout error on 40 trajectories of real-world deformable objects when three input frames are given. Fig. \ref{fig:real_world_quantitative} shows that TP-Net outperforms all other benchmarks in terms of both metrics and for all kinds of input ordering methods. It is not surprising that TP-Net and GraphRNN shows constant performance regardless of the input ordering method as they are invariant to input permutations. However, TP-Net outperforms GraphRNN by a large margin in both metrics, which demonstrates the effectiveness of our approach in processing a sequence of unordered point sets. For other models, the performance fluctuates depending on the input ordering method, but there exists no ordering that results in the best performance for all models, or results in higher performance than TP-Net. Although we have only tried three methods of input ordering, these results imply that there exists no ordering of a point set that ensures stable performance in the general sense, thus demonstrating the importance of achieving permutation invariance.

\subsection{Time and Space Complexity Analysis}
In this section, we compare the time and space complexity of TP-Net and GraphRNN \cite{gomes2021spatiotemporal} which are both capable of processing a sequence of unordered point sets for future prediction. 

Table \ref{complexity_table} shows the time complexity -- number of floating-point operations per sample (FLOPs/sample) -- and space complexity -- number of parameters in the network (\#params) -- of TP-Net and GraphRNN. While GraphRNN is more space efficient than TP-Net in terms of \#params in the network ($18.75$ times less parameters), TP-Net is significantly more efficient in computational cost as it has $354.4$ times less floating point operations per sample. As described in Section \ref{sec:consuming_unordered}, the time complexity of GraphRNN is $O(N^2)$ -- quadratic in the number of points -- because it searches for spatio-temporal neighbor points for every point in the current frame. However, the time and space complexity of TP-Net is $O(N)$ since it simply extracts global features from each point set by max pooling per point features, and reasons on those global features sequentially. This makes our approach more scalable than existing methods.
Empirically, TP-Net can predict more than 120 future point sets per second when tested on our real-world dataset using a GTX 1080Ti GPU on Tensorflow. Based on these results, we believe our framework can be widely used in real-time applications.

\begin{table}
\renewcommand{\arraystretch}{1.1}
\centering
\caption{Time and space complexity of deep architectures for processing a sequence of three unordered point sets with 30 particles. The ``M" stands for million.}
\label{complexity_table}
\begin{tabular}{ | m{6em} | m{2cm}| m{1.2cm} | }
  \hline
    & FLOPs/sample & \#params \\ 
  \hline
  TP-Net (ours) & 21.4M & 7.5M \\ 
  GraphRNN & 7584.6M & 0.4M \\ 
  \hline
\end{tabular}
\end{table}

\section{Conclusion}
We presented a deep neural network named TP-Net that can directly consume a sequence of unordered point sets to infer the future dynamics of a deformable object. By using global features of each point set, our model achieves invariance to $(N!)^m$ input permutations and shows state-of-the-art performance, even in long term prediction. We also demonstrated that our model well generalizes to real-world scenarios and is much more efficient than existing methods of consuming a sequence of unordered point sets. 

Although we have only used a 2D dataset in this study, in principle, our framework easily generalizes to datasets with higher dimensions. Moreover, we focused on consuming a sequence of static point sets in this study whereas many situations involve reasoning on dynamic point sets, where the number of points change over time. We believe that further works that address these concerns will empower and extend our framework with much more flexibility.

\section*{Acknowledgements}
This research was supported by the MSIT(Ministry of Science and ICT), Korea, under the ITRC(Information Technology Research Center) support program(IITP-2021-2018-0-01419) supervised by the IITP(Institute for Information and Communications Technology Planning and Evaluation) and the National Research Foundation of Korea(NRF) grant funded by the Korea government(MSIT). (No. NRF-2020R1A2C2014622)

\bibliography{main}

\appendix
\renewcommand{\thesection}{\Alph{section}.\arabic{section}}
\setcounter{section}{0}

\appendix
\renewcommand{\thesection}{\Alph{section}.\arabic{section}}
\setcounter{section}{0}

\newpage
\begin{appendices}

\section{Data Generation}
\label{sec:data_generation}
\subsection{Synthetic Data}
\subsubsection{Formulating a Deformable Object}

\begin{figure}[ht]
    \centering
    \begin{subfigure}{0.33\textwidth}
        \centering
        \includegraphics[width=\textwidth]{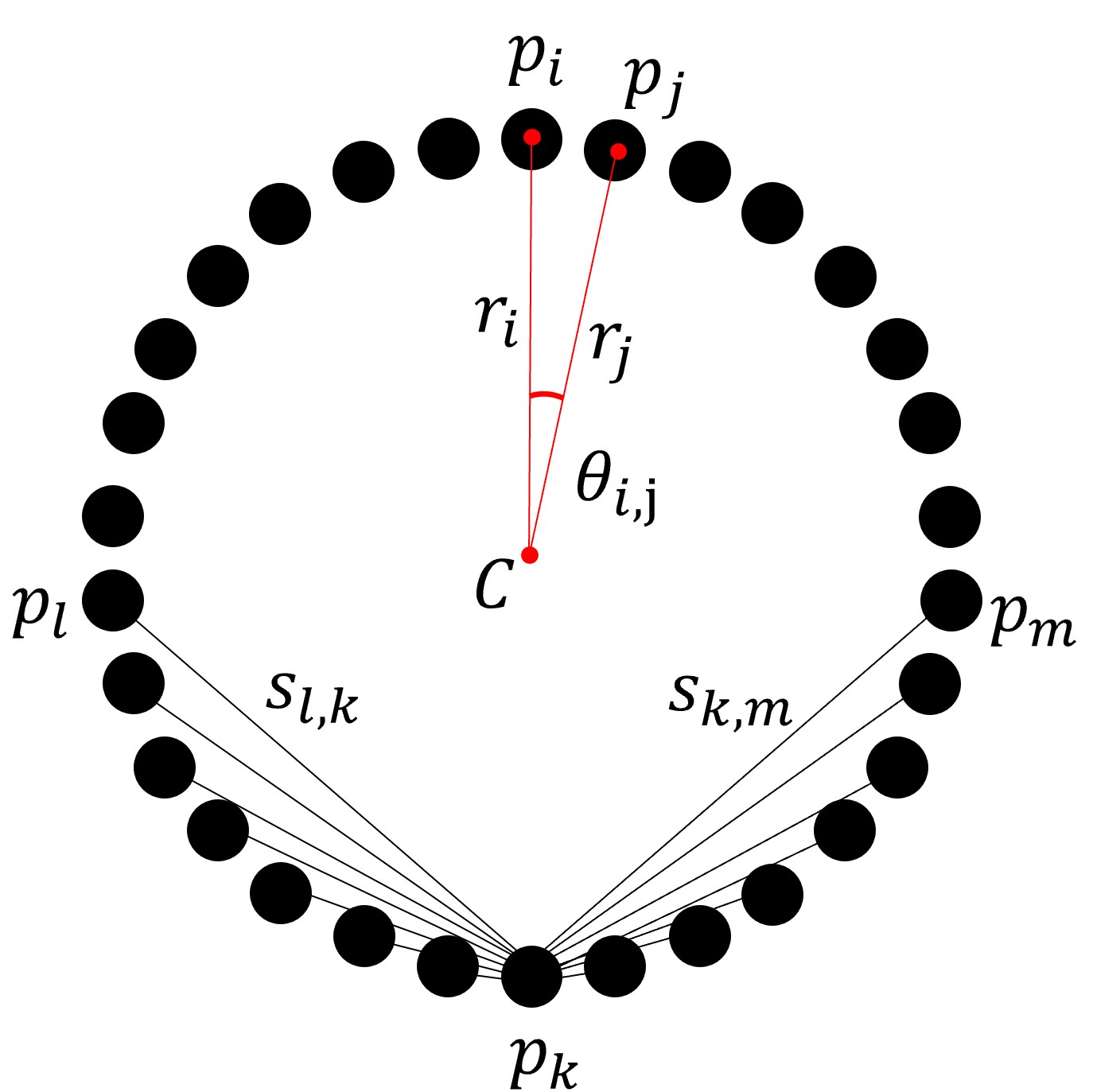}
        \caption{}
        \label{fig:softbody_rest}
    \end{subfigure}
    \hfill
    \begin{subfigure}{0.33\textwidth}
        \centering
        \includegraphics[width=\textwidth]{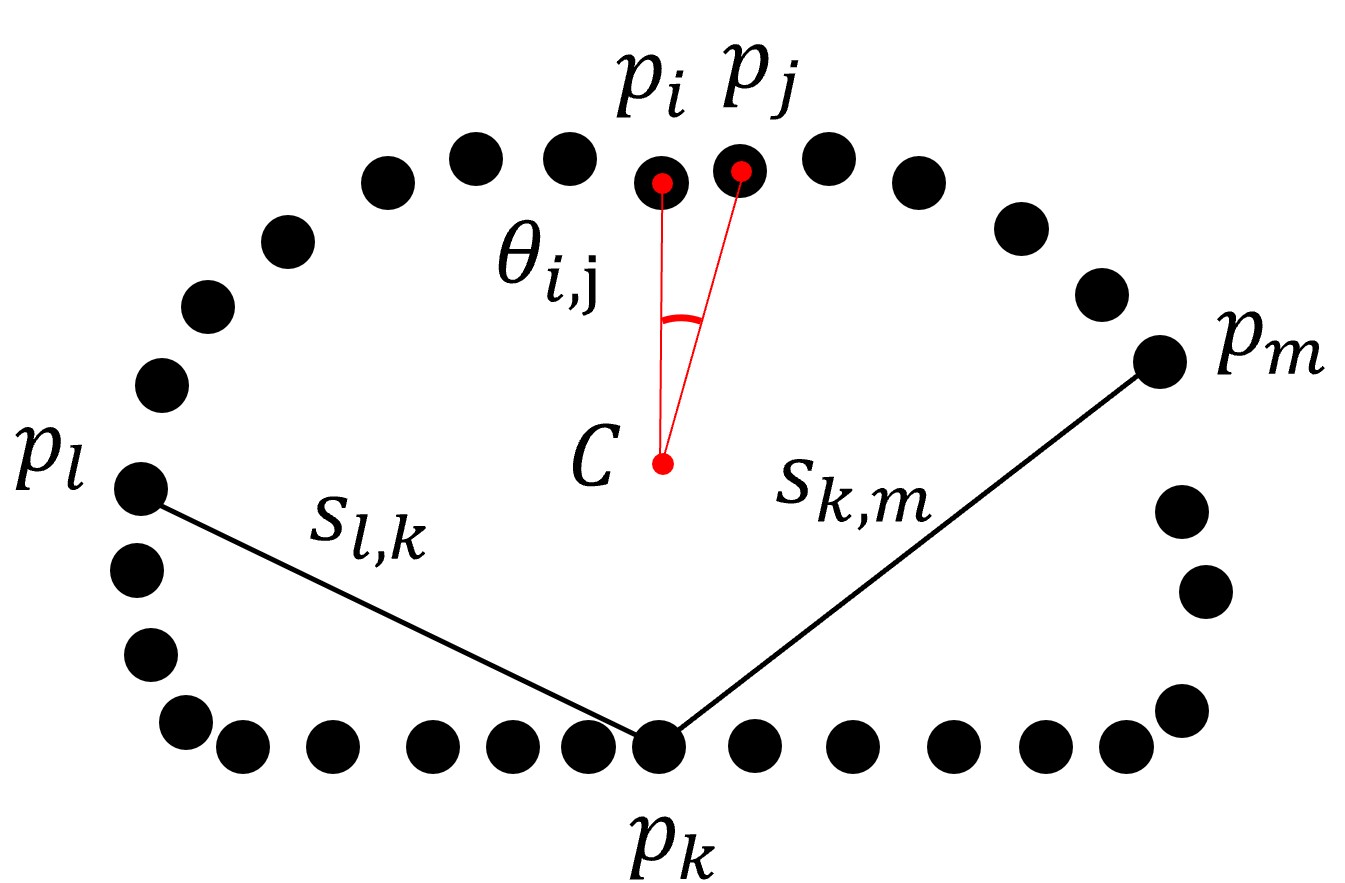}
        \caption{}
        \label{fig:softbody_deformed}
    \end{subfigure}
        \caption{A simplified diagram describing the formulation of a deformable object using the Box2D physics engine. (a) At rest state, the distance between the center $C$ of the object and each point $p_i$ is set to two -- i.e., $r_i = 2$ for all $i = 1, \ ... \ , n$. We also connect $p_l$ and $p_k$ with a restorable spring $s_{l,k}$ for all $| l - k | < \frac{2}{5}n$, although most of the spring connections that exist in the deformable object are omitted in this figure for simplicity. We also set $\theta_{i, j} = \frac{2\pi}{n}$ for all $i, j$ $\in \{1, \ ... \ , n\}$, which denotes the angle between neighboring points with respect to the center of the object. (b) Spring connections between particles enable deformation of the object.}
        \label{fig:softbody}
\end{figure}

\begin{figure*}[ht]
\centering
\includegraphics[width=0.99\textwidth]{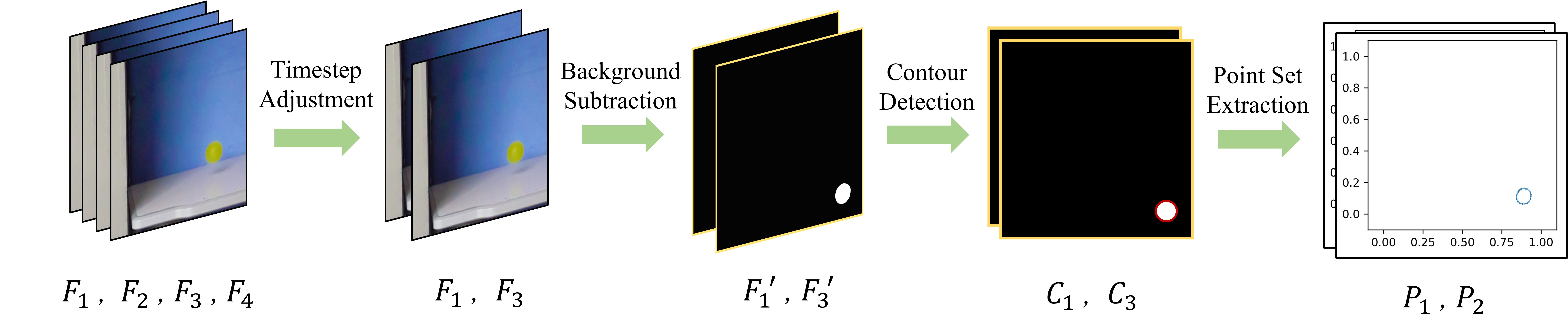}
\caption{\label{fig:real_world_preprocess}The process of obtaining point set data from raw video frames. $F_t$ denotes a frame captured from the original video at time $t$. $F_t^{'}$ denotes a frame with background subtraction at time $t$. $C_t$ denotes the contour of the object extracted from the frame at time $t$. Note that $P_t$ denotes a point set obtained from the original video frame at time $2t - 1$ due to time step adjustment.}
\end{figure*}

In this section, we describe how we formulate a deformable object when generating the synthetic dataset with Box2D physics engine \cite{catto2020box}. As illustrated in Fig. \ref{fig:softbody}, we model a deformable object with mass-spring system, that has $n$ particles $P = \{ p_{1}, p_{2}, ... , p_{n} \} $ comprising its boundary. For every particle pair $(p_{i}, p_{j})$ that is $| i - j | < \frac{2}{5}n$, we connect the particles with a restorable spring that shares the same spring constant across all springs within the object. In order to mimic the movement of a real-world deformable object that we use to generate our real-world dataset, we configure the Box2D environment as follows:
\begin{itemize}
    \item n = 30,
    \item gravity = (0, -0.5),
    \item friction of each particle = 1,
    \item friction of the wall = 1,
    \item frequency of the spring = 1,
    \item damping ratio of the spring = 0,
    \item restitution of each particle = 0,
    \item restitution of the wall = 1.
\end{itemize}

\subsubsection{Data preparation}
After we model a deformable object in a 2D Euclidean space, we generate 7125 simulated trajectories of a deformable object such that each trajectory is a sequence of 600 point sets. When generating simulated trajectories, we randomize the initial position, initial force (magnitude), and initial direction of the object. The scope of randomizing each property is as follows:
\begin{itemize}
    \item initial center position = $\{(x, y) \ | \ 2.9 \leq x, y \leq 42.1 \}$.
    \item initial force (magnitude) = $\{1.0, 1.15, 1.3, 1.45, 1.6\}$,
    \item initial direction = $\{\ang{180}, \ang{185}, \ang{190}, \ ... \ , \ang{270} \}$,
\end{itemize}

After we generate 7125 simulated trajectories, we construct our training dataset and validation dataset by sampling subsequences from each trajectory. Each sub-sequence is a sequence of 11 or 12 or 13 point sets, depending on the number of input frames we use, which varies from three to five point sets. For 240 trajectories, we set the entire trajectory as the training scope, and for 6635 trajectories, we only sample subsequences from certain parts to construct our training dataset. The rest of the trajectories are used to generate the validation dataset and test dataset. In order to increase the prediction accuracy of our model especially in terms of shape deformation, for each trajectory, we sample 15 subsequences that involves the interval in which the object collides with the wall, and sample 5 subsequences from a normal interval within the trajectory. We also normalize the position data of each points to be between 0 and 1 when training, and denormalize the data when visualizing the results.

When generating our test dataset, we generate 60 simulated trajectories with a unique pair of initial position, initial force, and initial direction that the model was never trained on, in order to test the generalization ability of our model.

\subsection{Real-World Data}

\begin{figure}[ht]
\centering
\includegraphics[width=\columnwidth]{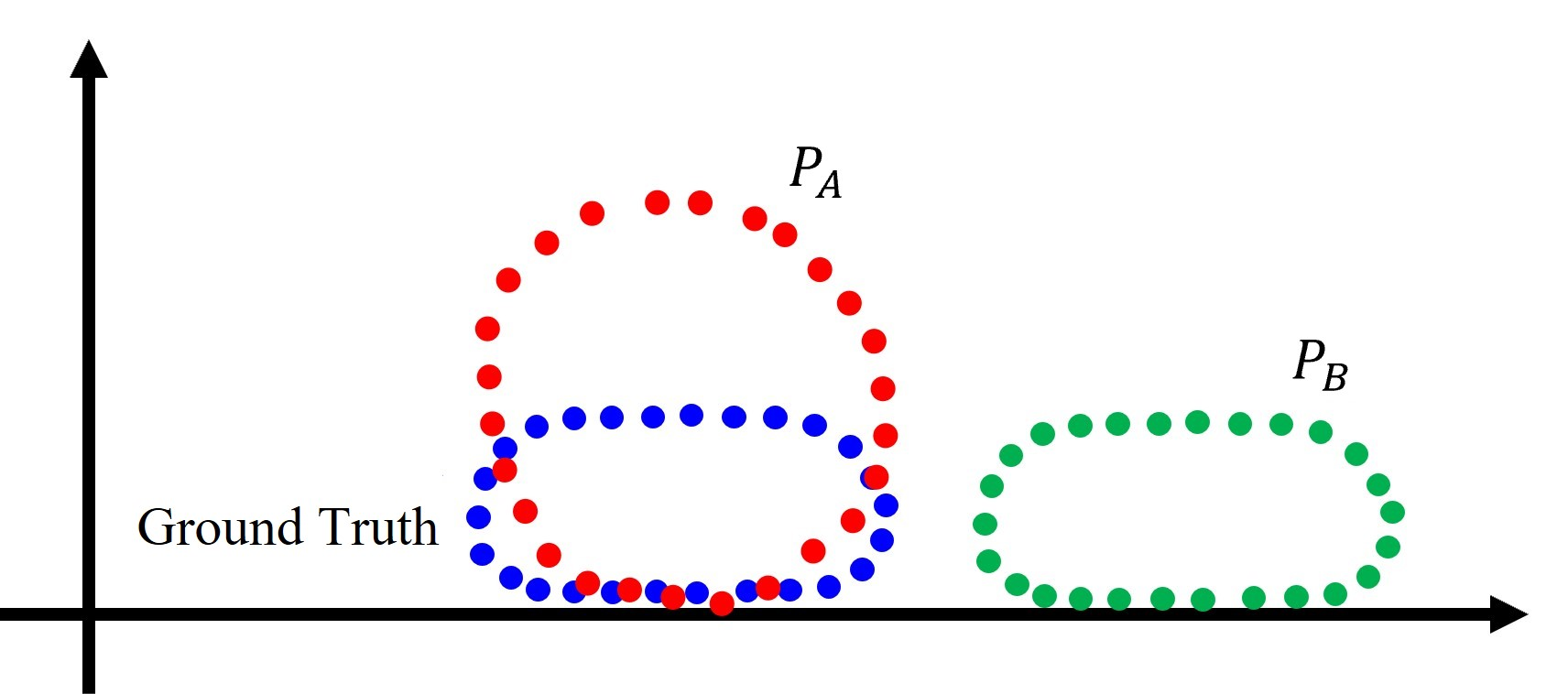}
\caption{Limitations of using the vanilla Chamfer distance (position error) as a single evaluation metric when comparing the performance of each model in our study.}
\label{fig:shape_error_reason}
\end{figure}

\begin{figure*}[ht]
\centering
\includegraphics[width=\textwidth]{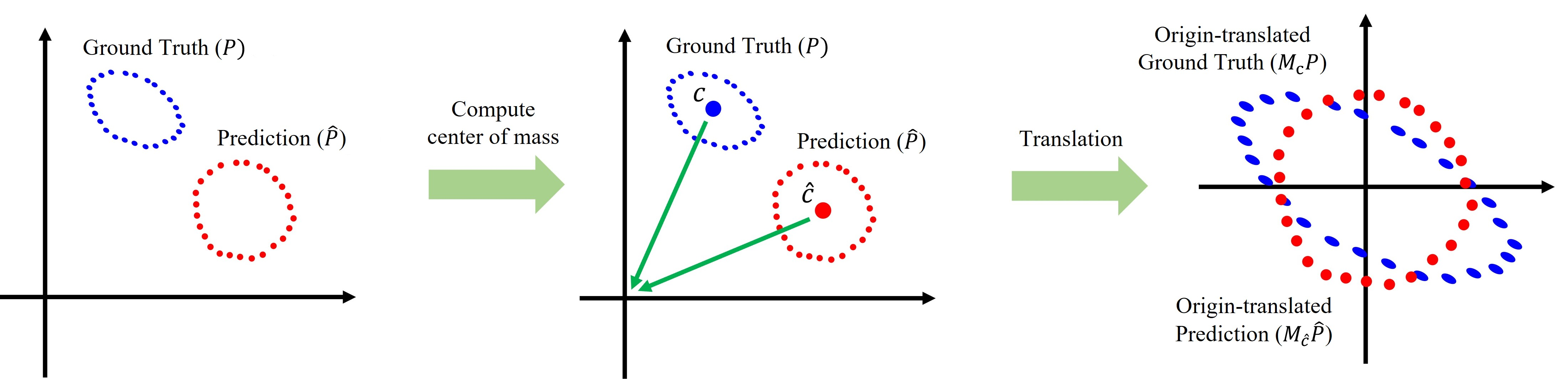}
\caption{The process of computing shape error $E_s$.}
\label{fig:shape_error_process}
\end{figure*}

\subsubsection{Generating Real-World Trajectories}
Due to the absence of public dataset for moving real-world deformable objects, we collect and construct a novel real-world dataset from scratch in our study. By using a high-speed camera that can capture up to 480 frames per second, we film a fast moving real-world deformable object. We throw a deformable object from various positions with varying forces and directions to be similar but not the same with the synthetic dataset. Since we convert our video data into a sequence of 2D point sets, we control the depth value of the object by throwing the object in parallel with the camera at a constant distance.

\subsubsection{Data preprocessing} 
We preprocess the video data to obtain the particle positions data that our model can consume as input. As illustrated in Fig. \ref{fig:real_world_preprocess}, we apply a series of operations to the original video frames to obtain the point set $P$ for each time step. First, we sample video frames at odd time steps to adjust the time step of real-world deformable object to be similar to that of the simulated deformable object which our model was trained on. Then, we subtract the background of each frame to identify the contour $C$ of the deformable object. After we obtain the contour of the object, we uniformly divide the contour into $n$ points as described in Fig. \ref{fig:softbody} to obtain the particle positions data $P$. Although the obtained point sets $P_1$ and $P_2$ in Fig. \ref{fig:real_world_preprocess} represent the states that are originally two time steps away from each other, we define them as point sets with consecutive time steps when composing our final real-world dataset. Note that we also normalize the position data to be between 0 and 1 by using the relative pixel coordinates with respect to the wall and floor. Conversely, when we visualize the predicted point sets, we plot the points on the background image by computing its relative pixel coordinate.

\section{Evaluation}
\subsubsection{Adoption of shape error} 
\label{sec:adoption_of_shape_error}
As described in Section \ref{sec:experiment}, we use two metrics to evaluate our model -- position error($E_p$) and shape error($E_s$). The purpose of using two metrics for evaluation is to evaluate the prediction accuracy of our model in two different perspectives. As shown in \eqref{position_error_equation}, position error is the Chamfer distance between the ground truth point set and predicted point set, which measures the distance between individual particles. In Fig. \ref{fig:shape_error_reason}, $P_A$ is a better prediction of the ground truth than $P_B$ in terms of position error since the distances between individual particles in each point set are much closer than that of $P_B$. However, it is clear that $P_B$ is a much better prediction of the ground truth in terms of shape. Therefore, using the vanilla Chamfer distance as a single evaluation metric can mislead to false interpretation of the performance of the model in these cases. In order to prevent such misunderstanding, we define our shape error as described in \eqref{shape_error_equation}.

\subsubsection{Computing shape error} In this section, we provide a detailed explanation of \eqref{shape_error_equation} which is used for computing the shape error. As illustrated in Fig. \ref{fig:shape_error_process}, we first compute the center of mass $c$ and $\hat{c}$ of each point set $P$ and $\hat{P}$, respectively. Then, we apply a translation matrix $M_c$ and $M_{\hat{c}}$ to $P$ and $\hat{P}$, respectively, to shift the center of each point set to the origin. After translating each point set, we then compute the Chamfer distance between these translated point sets to compute the shape error.

\end{appendices}
\end{document}